\documentclass[letterpaper, 10pt, journal, twoside]{IEEEtran}
\usepackage{amsmath,amsfonts}
\usepackage{algorithmic}
\usepackage{algorithm}
\usepackage{array}
\usepackage[caption=false,font=normalsize,labelfont=sf,textfont=sf]{subfig}
\usepackage{textcomp}
\usepackage{stfloats}
\usepackage{url}
\usepackage{verbatim}
\usepackage{graphicx}
\usepackage{cite}
\usepackage[hidelinks]{hyperref}
\usepackage{cleveref}
\usepackage[hidelinks]{hyperref}
\usepackage{comment}
\begin{document}

\title{Real-Time Shape Estimation of Tensegrity Structures Using Strut Inclination Angles
}

\author{
    Tufail Ahmad Bhat$^{1}$,
    Yuhei Yoshimitsu$^{1}$,
    Kazuki Wada$^{1}$,
    Shuhei Ikemoto$^{1}$
\thanks{Manuscript received: December 2, 2024; Revised: February 17, 2025; Accepted: March 25, 2025.}
\thanks{This paper was recommended for publication by Yong-Lae Park upon evaluation of the Associate Editor and Reviewers' comments.}
\thanks{This work was supported by JSPS KAKENHI Grant Number 23K24927.}
\thanks{$^{1}$All authors are with the Kyushu Institute of Technology, Japan (e-mail:  {\tt\small ikemoto@brain.kyutech.ac.jp})  }
\thanks{Digital Object Identifier (DOI): see top of this page.}
}

\markboth{IEEE Robotics and Automation Letters. Preprint Version. Accepted March, 2025}
{BHAT \MakeLowercase{\textit{et al.}}: TENSEGRITY ROBOT SHAPE ESTIMATION}


\maketitle

\begin{abstract}
Tensegrity structures are becoming widely used in robotics, such as continuously bending soft manipulators and mobile robots to explore unknown and uneven environments dynamically. Estimating their shape, which is the foundation of their state, is essential for establishing control. However, on-board sensor-based shape estimation remains difficult despite its importance, because tensegrity structures lack well-defined joint structures, which makes it challenging to use conventional angle sensors such as potentiometers or encoders for shape estimation. To our knowledge, no existing work has successfully achieved shape estimation using only onboard sensors such as Inertial Measurement Units (IMUs). This study addresses this issue by proposing a novel approach that uses energy minimization to estimate the shape. We validated our method through experiments on a simple Class 1 tensegrity structure, and the results show that the proposed algorithm can estimate the real-time shape of the structure using onboard sensors, even in the presence of external disturbances.

\end{abstract}

\begin{IEEEkeywords}
Tensegrity Robots, Inertial Measurement Unit (IMU), Onboard Sensors, Real-Time Shape Estimation
\end{IEEEkeywords}

\section{Introduction}
\begin{figure}[!t]
    \centering
    \includegraphics[width=88mm,height=49mm]{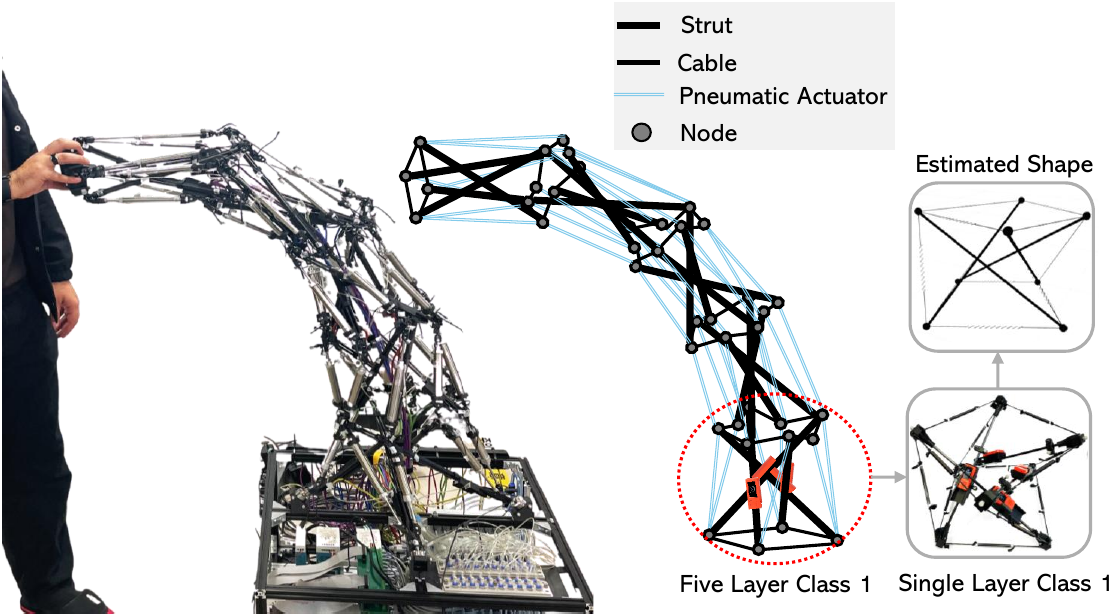}

    \caption{\textbf{Left} TM40: Tensegrity manipulator consisting of 20 struts, 40 pneumatic actuators, and 40 cables. \textbf{Middle}: Schematic of TM40. Unlike articulated robots with well-defined joints, this manipulator has no joints. \textbf{Right}: The estimated shape of a simple Class 1 tensegrity.
    }
    \label{fig:TM40 tensegrity manipulator}
 
\end{figure}

\IEEEPARstart{T}{he} concept of “tensegrity” is coined by the iconoclastic architect R. Buckminster Fuller. It describes structures that achieve stability through a balance of forces: specific components, known as “cables" are always in tension, while others, known as “struts" are constantly under compression \cite{skelton2009tensegrity}. 
In tensegrity, the cables of the structure are always under continuous tension, a condition known as “prestress”. Thanks to its nature, where the shape is determined by the mechanical equilibrium, tensegrity is inherently flexible and highly resistant to impact. Due to these advantages, tensegrity structures have been widely studied for applications in mobile robots \cite{mintchev2018soft,kim2014rapid,kim2016hopping,kimber2019low,jeong2024spikebot}, joint structures \cite{lessard2016lightweight,wang2024twrist}, and manipulators \cite{lessard2016bio,ramadoss2022hedra,kobayashi2023large}. For instance, our research group has been developing the tensegrity manipulator shown in Fig.~\ref{fig:TM40 tensegrity manipulator}, as a platform for motor learning in long-term interactive scenarios \cite{yoshimitsu2022development}.

In traditional articulated robots, the joint angles primarily define the configuration, if these joint angles are accurate, we can use these measurements in the kinematic equations to calculate the robot's posture without significant errors. The articulated robot links do not deform under normal conditions, and clear joints make it easier to determine their configuration.
 In contrast to traditional articulated robots, tensegrity robots lack clear joints and rigidity, which makes it difficult to measure their configurations, that is, their shapes \cite{shah2022tensegrity} using conventional angle sensors. To overcome this issue, alternative sensors must be embedded into the structure. If the tensegrity robot is driven by electric motors via cables, the cable length becomes important information. Theoretically, if the shape of the rigid body is known, all cable lengths completely define the shape.  However, to date, tensegrity robots that actively and independently control all cables have not been developed, and this missing information is typically compensated using additional sensors. For instance, \cite{johnson2022sensor} introduced capacitive tendon sensors to measure the lengths of undriven cables. This approach aims to determine the shape based on the lengths of all members, however, its broader application remains a topic for future research.

An Inertial Measurement Unit (IMU) is a promising sensor for addressing the challenge of missing information in shape estimation. For instance, \cite{huang2022live} presented a live demonstration of the first three-bar tensegrity robot that is capable of estimating its shape relative to a global frame using nine passive tendon sensors and two IMUs. Similarly,  \cite{tong2025tensegrity} employed IMUs and motor encoder sensors to estimate the shape and pose. Likewise, \cite{caluwaerts2016state} combined local shape estimation with global pose estimation using both internal (IMUs, motor encoders) and external ultra-wideband ranging sensors. When an underdriven mechanism is used, measuring the length of each cable is challenging. As a result, shape estimation methods that do not rely on cable length information have also been proposed. Visual sensors have been extensively studied as an alternative. For example, \cite{moldagalieva2019computer} proposed a vision-based approach for shape estimation in tensegrity robots. In addition, \cite{bezawada2022shape} introduced a shape-estimation algorithm that integrates IMUs with visual tracking.

As discussed above, various methods have been explored for the shape estimation of tensegrity robots. Unlike conventional articulated robots, which typically rely on a single type of sensor for configuration measurement, regardless of the drive system or application, these methods often require multiple sensors, making it difficult to generalize across different drive systems and applications. For specific robots or tasks, it is possible to define and estimate the state \cite{huang2022live, booth2021surface, lu20226n} with an effort comparable to that required for the shape estimation. Many of the aforementioned studies performed state estimation simultaneously with shape estimation. However, considering that the shape of tensegrity robots is equivalent to the configuration of conventional articulated robots, it is desirable to address shape estimation in a generalized problem setting with minimal sensor requirements, even if it requires prior knowledge of the detailed design parameters.

To address the aforementioned issues, we propose a method that estimates the shape, and our main contributions can be summarized as follows :
\begin{itemize}
    \item We propose a shape estimation method for
tensegrity structures that depends solely on the inclination angle of each strut.

 \item We formulate a gradient-based method to estimate the orientation of a strut using only inclination data obtained from 6-axis IMUs, without relying on a magnetometer sensor.

 \item We demonstrate the algorithm's real-time implementation for shape estimation by solving the energy function using a gradient-based approach, which uses only inclination angle information.
\end{itemize}

The validity of the proposed method has been verified using the four-strut tensegrity prism setup as shown in Fig.~\ref{fig:class1}. The proposed method assumes that
the connectivity matrix, which defines the structure of the
tensegrity robot, and the linear stiffness coefficients of the cables are known or identified as the design parameters.

The structure of the paper is as follows. The proposed real-time shape reconstruction algorithm is described in \hyperref[sec:ProposedAlgorithm]{Section II}. The details of the experimental setup, implementation, and experimental results appear in \hyperref[sec:experimental]{Section III}. Finally, the conclusions and directions for future work are summarized in \hyperref[sec:conclusion]{Section VI}.
\section{Proposed Algorithm}
\label{sec:ProposedAlgorithm}
\subsection{Notation and Nomenclature}    
The following notations and definitions are used throughout this paper. The definitions of matrices, vectors, and elements follow the conventions described by \cite{petersen2008matrix}.
\begin{itemize}
    \item \textbf{{A}}: $m \times n$ matrix, where $m$ is the number of rows and $n$ is the number of columns.
    \item \(\mathbf{a} \): A row vector in $\mathbb{R}^{n}$.
    \item $a$: A scalar in $\mathbb{R}$.
    \item $a_i$: The $i$\textsuperscript{th} element of the vector $\mathbf{a}$, with $i \in \{1, \dots, n\}$.

\end{itemize}
In this study, we use the terms “shape reconstruction” and “shape estimation” interchangeably. In addition, we refer to “Onboard + External” as “Onb/Ex”.

\subsection{Kinematic Relationship}
Our algorithm estimates the spatial positions of the nodes connected by each strut. To achieve this, we decompose the nodes of the struts into two key components, along with the length of the strut: 
the center position and the orientation of each strut are denoted by \(\mathbf{p}_i\in \mathbb{R}^3 \) and \(\mathbf{q}_i\in \mathbb{R}^3 \) respectively, where \( i \) ranges from 1 to \( m_b \), with 
\(m_b \)
  is the total number of struts in the structure, and \(\mathbf{q}_i\) is a unit vector ($\|\mathbf{q}_i\|$ = 1) parallel to the strut's longitudinal direction, with \(\mathbf{p}_i \) as its origin.

\begin{figure}[!t]
    \centering
    \includegraphics[height=32mm]{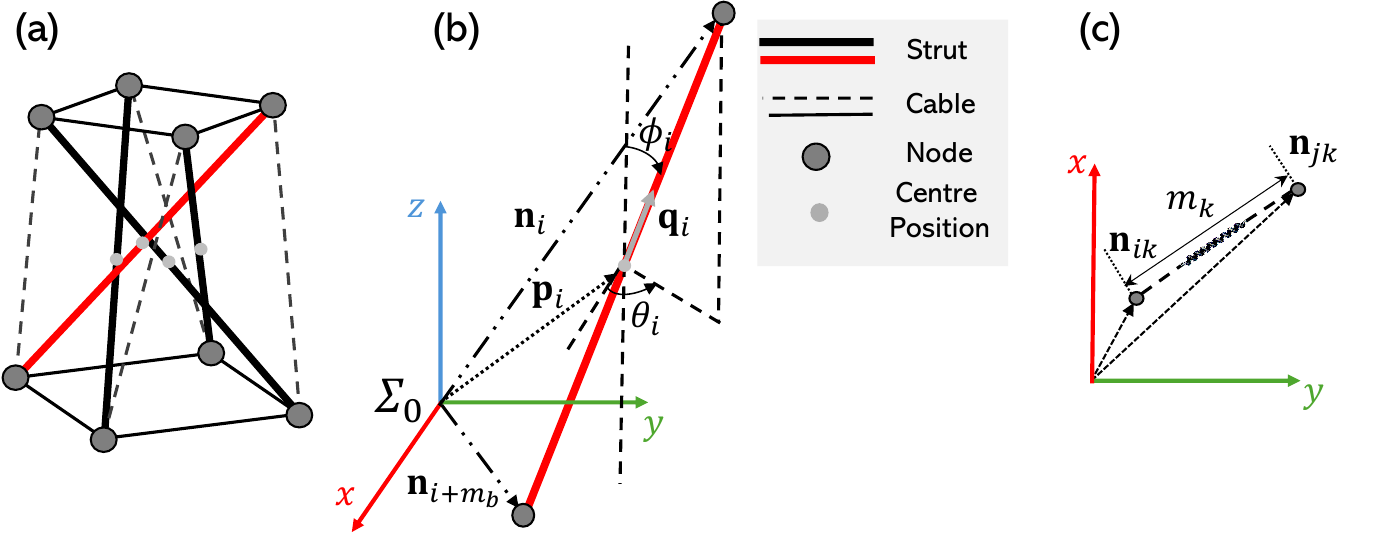}

    \caption{
        {Geometric Representation of a Tensegrity Structure}
        (a) Overview of the Class 1 tensegrity structure. 
        (b) Geometric representation of \(i\)\textsuperscript{th} strut element in the \(xyz\)-coordinate frame and the relevant angles, such as the inclination angle \(\phi_i\) and the rotational angle \(\theta_i\) around the gravity vector or yaw angle.
        (c) Schematic representation of \(k\)\textsuperscript{th} cable element, connecting nodes \(\mathbf{n}_{ik}\)
and \(\mathbf{n}_{jk}\) in a two-dimensional plane.
    }
    \label{fig:geometric}
    
\end{figure}

For strut \( i \) shown in Fig.~\ref{fig:geometric}(b), the spatial positions of the two connected nodes are calculated as follows :

  \begin{equation}
\begin{aligned}
\mathbf{n}_i & = \mathbf{p}_i+ \frac{L_i}{2} \mathbf{q}_i, & 
\mathbf{n}_{i + m_b} & = \mathbf{p}_i - \frac{L_i}{2} \mathbf{q}_i
\end{aligned}
\label{eq:two_nodes}
\end{equation}

where \( \mathbf{n}_i \) and  \(\mathbf{n}_{i + m_b}\) represent the positions of the two nodes connected to the strut \( i \), and \( L_i \) is the length of the strut. Specifically, if the rotation around the longitudinal axis of the strut is ignored, vector $\mathbf{q}_i$ is defined as follows :

\begin{equation}
\begin{gathered}
\mathbf{q}_i=f(\phi_i, \theta_i) \\
\end{gathered}
\label{eq:q}
\end{equation}

where $\phi_i \in \mathbb{R}$ represents the angle of inclination relative to the gravity vector, and $\theta_i \in \mathbb{R}$
 is a yaw or rotational angle around the gravity vector. 

\subsection{Parameter Identification: Known and Unknown Variables}
Here, we analyze the variables in (\ref{eq:two_nodes}) to identify the known and unknown terms. The variable $\mathbf{p}_i$ is the absolute position of the strut in three-dimensions, and direct measurement of this position is challenging because the IMU sensor does not provide an absolute position. Thanks to that the energy function in the proposed method depends on relative positions, this difficulty does not make the shape estimation problem challenging. 
On the other hand, the length $L_i$, is known and remains constant. To obtain the orientation $\mathbf{q}_i$ of the strut, we require two angle measurements: $\phi_i$ and $\theta_i$. Traditionally, we determine $\phi_i$ using a 3-axis accelerometer and $\theta_i$ using 3-axis magnetometer sensors. However, using multiple magnetometers in close proximity is impractical because of magnetic interference, therefore, the strut orientation is only partially known. We discuss the challenges with IMU sensors in detail in the  \hyperref[subsec:challenges]{subsection H}.

\subsection{Matrix and Vector Representation}
Let \(\mathbf{n} \in \mathbb{R}^{3n}\) represent the concatenated vector of positions for each node, where \(n\) is the total number of nodes in the structure. The vector \(\mathbf{n}\) concatenates all the top and bottom nodes, which can be expressed as :
\begin{equation}
\begin{aligned}
\mathbf{n} = \begin{bmatrix} n_{1x} & n_{1y} & n_{1z} & \dots & n_{nx} & n_{ny} & n_{nz} \end{bmatrix}^T \in \mathbb{R}^{3n \times 1}
\end{aligned}
\label{eq:me}
\end{equation}
 Similarly, let $\mathbf{p} \in \mathbb{R}^{3m_b}$ represent the concatenated vector of the center positions for each strut with  \(m_b\) as the total number of struts, which can also be expressed as \(m_b = \frac{n}{2}\). The vector $\mathbf{p}$ is defined as :
\begin{equation}
\begin{aligned}
\mathbf{p} = \begin{bmatrix} p_{1x} & p_{1y} & p_{1z} & \dots & p_{m_bx} & p_{m_by} & p_{m_bz} \end{bmatrix}^T \in \mathbb{R}^{3m_b \times 1}
\end{aligned}
\label{eq:u}
\end{equation}
The total number of the elements in the structure is \(m\), and it can be written as \(m =m_b+m_s
\), where \(m_s\) is the total number of cables. Similarly, \(\mathbf{q} \in \mathbb{R}^{3m_b}\) represents the concatenated vector of the orientation of each strut. Thus, the vector \(\mathbf{q}\) can be written as :
\begin{equation}
\begin{aligned}
\mathbf{q} = \begin{bmatrix} q_{1x} & q_{1y} & q_{1z} & \dots & q_{m_bx} & q_{m_by} & q_{m_bz} \end{bmatrix}^T \in \mathbb{R}^{3m_b \times 1}
\end{aligned}
\label{eq:ut}
\end{equation}

In general, we can write the relationships from (\ref{eq:two_nodes}) using the vectors $\mathbf{n}$, $\mathbf{p}$, and $\mathbf{q}$ as follows :
\begin{equation}
\mathbf{n} = \mathbf{A}^T \mathbf{p} + \mathbf{B}^T \mathbf{q} \in \mathbb{R}^{3{n}}
\label{eq:nodalvector}
\end{equation}
where
\[
\mathbf{A} = \left[ \mathbf{I}_{3m_b} \mid \mathbf{I}_{3m_b} \right], \quad \mathbf{B} = \left[ \frac{\mathbf{L}}{2} \mid -\frac{\mathbf{L}}{2} \right]
\]

Here, $\mathbf{I}_{3m_b}$ denotes an identity matrix of size \( 3m_b \) and $\mathbf{L}$ is a diagonal matrix defined as \( \operatorname{diag}(L_1 \cdots L_{3m_b}) \).

\subsection{Elastic Energy Formulation}
 A tensegrity structure consists of cables and strut elements, and the points of intersection between these elements are known as “nodes". The forces at these nodes are balanced, which corresponds to the structure being in a minimum energy configuration \cite{skelton2009tensegrity}, thus the problem of determining $\mathbf{p}_i$ and $\theta_i$ in the structure can be reduced to an energy-minimization problem. The total elastic energy \( e \) stored in all cable elements is the sum of the individual cable energies, $e_k$, and it can be expressed as :
 \begin{equation}
\label{eq:deqn_ex2}
e = \sum_{k=1}^{m_s} e_k, \quad
e_k =  \frac{1}{2} K_k \left( m_k - b_k \right)^2
\end{equation}

where \( K_k \), \( m_k \), and \(b_k\) correspond to the stiffness, current length, and the natural length of the $k$\textsuperscript{th} cable, respectively as shown in Fig.~\ref{fig:geometric}(c). For simplicity, we assume that the cable is initially at its natural length \( b_k = 0 \). The length \( m_{k} \) of the cable between nodes \( \mathbf{n}_{ik} \) and \( \mathbf{n}_{jk} \) can be expressed as the Euclidean distance between node positions.

\begin{equation}
\begin{aligned}
m_k=\left\|{\mathbf{n}_{ik}}-  \mathbf{n}_{jk}\right\|=\left\|\mathbf{n} 
\mathbf { d}_k\right\|
\end{aligned}
\label{eq:r}
\end{equation}
 where \(\mathbf { d}_k \) is a 3\( n \)-dimensional vector representing the connectivity between nodes, with a value of 1 at \( \mathbf{n}_{ik} \) node, -1 at \( \mathbf{n}_{jk} \) node, and 0 otherwise.

In the general form, the total elastic energy \( e\) stored in all cables in (\ref{eq:deqn_ex2}), can be expressed as :
\begin{equation}
\begin{aligned}
e = \frac{1}{2}  \mathbf {m_s}^T \mathbf {K} \mathbf {m_s}
\end{aligned}
\label{eq:energy_in_summation}
\end{equation}
Equation (\ref{eq:energy_in_summation}) represents the quadratic form of the elastic energy, where \( \mathbf{K} = \operatorname{diag}(K_1, \dots, K_{3m_s}) \) represents the stiffness matrix, and $\mathbf {m_s}$ represents the lengths of all cables and is defined as :
\begin{equation}
\begin{aligned}
& \mathbf {m_s} = \mathbf {C_s} \left( \mathbf {A} ^T \mathbf {p} + \mathbf {B} ^T \mathbf {q}  \right)
\end{aligned}
\label{eq:ms}
\end{equation}

Here, $\mathbf{C_s}$ represents the connectivity matrix of the cables.
\subsection{Connectivity Matrix}
The connectivity matrix $\mathbf{C}$ consists of values 1, 0, and -1, with each row indicating how the elements in the structure are connected to nodes. It can be written as :
\begin{equation}
\begin{aligned}
\mathbf{C} & =\mathbf{D}^T=\left[{\mathbf d}_1 \cdots {\mathbf d}   _{3 m}\right]^T \in \mathbb{R}^{3 m \times 3 n} \\  
\end{aligned}
\label{eq:C}
\end{equation}
Also, the connectivity matrix $\mathbf{C} $ in (\ref{eq:C}) can be written as :

\begin{equation}
\label{eq:whole_c}
\mathbf{C} =\left[\frac{\mathbf{C_b}}{\mathbf{C_s}}\right] \in \mathbb{R}^{3(m_b+m_s) \times 3 n}
\end{equation}
where
\begin{align}
\mathbf{C_b} & =\mathbf{D_b}^T=\left[{\mathbf d}_1 \cdots {\mathbf d}_{3 m_b}\right]^T \in \mathbb{R}^{3 m_b \times 3 n} \\
\mathbf{C_s} & =\mathbf{D_s}^T=\left[{\mathbf d}_1 \cdots {\mathbf d}_{3 m_s}\right]^T \in \mathbb{R}^{3 m_s \times 3 n}
\end{align}
\label{eq:whole_cc}

Equation (\ref{eq:whole_c}) is a concatenation of $\mathbf{C_b}$, which is the connectivity matrix of strut elements, and  $\mathbf{C_s}$ the connectivity matrix of the cable elements to form a full connectivity matrix.

For element \( k \) shown in Fig.~\ref{fig:geometric}(c), if it connects node \( \mathbf{n}_{ik} \) with node \( \mathbf{n}_{jk} \), the corresponding row in matrix  $\mathbf{C_s}$ will have a +1 in the column for node \( \mathbf{n}_{ik} \), a -1 in the column for node \( \mathbf{n}_{jk} \), and 0 otherwise, and the sum of the values in each row is always zero.
Each entry of $\mathbf{C_s}(u,v)$ is defined by the following rule :
\begin{equation}
\label{deqn_ex4}
\mathbf{C_s}(u,v) = 
\begin{cases} 
      +1 & \text{if } v = \mathbf{n}_{ik}, \\ 
      -1 & \text{if } v = \mathbf{n}_{jk}, \\ 
      0  & \text{otherwise}.
\end{cases}
\end{equation}
Here, \( u \) represents the row index for a cable, \( v \) is the column index corresponding to each node, and \( \mathbf{n}_{ik} \) and \( \mathbf{n}_{jk} \) are the indices of the two nodes connected by element \( k \).
In the case of our Class 1 tensegrity structure, the numbering of the nodes is shown in Fig.~\ref{fig:class1}, and the corresponding one-dimensional connectivity matrix  \( (\mathbf{C_s})_{x}\) is obtained as follows :
\[
\renewcommand{\arraystretch}{0.6} 
\setlength{\tabcolsep}{.6pt} 
(\mathbf{C_s})_{x} =
\begin{pmatrix}
1  & -1 & 0  & 0  & 0  & 0  & 0  & 0 \\
0  & 1  & -1 & 0  & 0  & 0  & 0  & 0 \\
0  & 0  & 1  & -1 & 0  & 0  & 0  & 0 \\
-1 & 0  & 0  & 1  & 0  & 0  & 0  & 0 \\
0  & 0  & 0  & 0  & 1  & -1 & 0  & 0 \\
0  & 0  & 0  & 0  & 0  & 1  & -1 & 0 \\
0  & 0  & 0  & 0  & 0  & 0  & 1  & -1 \\
0  & 0  & 0  & 0  & -1 & 0  & 0  & 1 \\
1  & 0  & 0  & 0  & 0  & -1 & 0  & 0 \\
0  & 1  & 0  & 0  & 0  & 0  & -1 & 0 \\
0  & 0  & 1  & 0  & 0  & 0  & 0  & -1 \\
0  & 0  & 0  & 1  & -1 & 0  & 0  & 0 \\
\end{pmatrix}
\begin{array}{l}

\left. \rule{0pt}{1.7em} \right\} 
\vcenter{\hbox{\rotatebox{90}{\scalebox{0.5}{\text{Upper-Loop}}}}} \\[1.5em]
\left. \rule{0pt}{1.7em} \right\} 
\vcenter{\hbox{\rotatebox{90}{\scalebox{0.5}{\text{Bottom-Loop}}}}} \\[1.5em]
\left. \rule{0pt}{1.7em} \right\} 
\vcenter{\hbox{\rotatebox{90}{\scalebox{0.5}{\text{ Interconnected }}}}}
\end{array}
\]
The rows represent the number of connections between nodes (12 connections in total), and the columns represent the number of nodes (8 nodes in total). This matrix captures the connectivity between the nodes and cable elements.

\subsection{Objective Function for Shape Reconstruction}
The quadratic form of the elastic energy in (\ref{eq:energy_in_summation}) can also be expressed as :
\begin{equation}
\begin{aligned}
e & = \frac{1}{2} \left( \mathbf{C_s} \mathbf{A}^T \mathbf{p}+ \mathbf{C_s} \mathbf{B}^T \mathbf{q} \right)^T \mathbf{K} \left( \mathbf{C_s} \mathbf{A}^T \mathbf{p} + \mathbf{C_s} \mathbf{B}^T \mathbf{q} \right)
\end{aligned}
\label{eq:e_elongation}
\end{equation}
 We formulate (\ref{eq:e_elongation}) as an optimization problem with the goal of obtaining the estimated \(\mathbf{\tilde{p}} \), \(\boldsymbol{\tilde{\theta}}\), and \(\mathbf{\tilde{n}} \) by minimizing the total energy $e$.
\begin{equation}
\mathbf{\tilde{p}} , \boldsymbol{\tilde{\theta}}\ = \underset{\boldsymbol{p}, \boldsymbol{\theta}}{\operatorname{argmin}} \, e
\label{eq:arg_min}
\end{equation}
\[
\begin{gathered}
\mathbf{\tilde{n}} \gets \mathbf{A}^T \mathbf{\tilde{p}} + \mathbf{B}^T \mathbf{q(\boldsymbol{\phi},\boldsymbol{\tilde{\theta}})} 
\end{gathered}
\]

Once we obtain the estimated spatial node vector \(\mathbf{\tilde{n}} \), the estimated shape can be reconstructed.
\begin{algorithm}[t]
\caption{Real-time Shape Reconstruction}
\label{alg:param_estimation_orientation}
\begin{algorithmic}[1]
    \STATE \textbf{Input:} Initial parameters $\mathbf{p}_0$, $\boldsymbol{\theta}_0$, number of steps $N = 300$, learning rates $\alpha = 0.0001$, $\beta = 0.0005$, max iterations for $\boldsymbol{\theta}$ and $\mathbf{p}$: $N_{\theta} = 1$, $N_{p} = 1$
    \STATE \textbf{Output:} Estimated parameters $\mathbf{p}$, $\boldsymbol{\theta}$, and $\mathbf{n}$
    \STATE \textbf{Initialize:} $\mathbf {p} \gets \mathbf {p}_0$, $\boldsymbol{\theta} \gets \boldsymbol{\theta}_0$   
    \FOR{$i = 1$ \textbf{to} $N$}
        \STATE Compute energy: $e \gets g(\mathbf{p}, \boldsymbol{\theta})$
        \STATE Compute orientation vector: $\mathbf{q} \gets f(\boldsymbol{\phi}, \boldsymbol{\theta})$
        
        \FOR{$j = 1$ \textbf{to} $N_{\theta}$}
            \STATE Compute gradients for $\boldsymbol{\theta}$: $\nabla_{\boldsymbol{\theta}} e$
            \STATE Update parameters: $\boldsymbol{\theta} \gets \boldsymbol{\theta} - \alpha \nabla_{\boldsymbol{\theta}} e$
        \ENDFOR
        
        \FOR{$k = 1$ \textbf{to} $N_{p}$}
            \STATE Compute gradients for $\mathbf{p}$: $\nabla_{\mathbf{p}} e$
            \STATE Update position: $\mathbf{p} \gets \mathbf{p} - \beta \nabla_{\mathbf{p}} e$
        \ENDFOR
    \ENDFOR
    \STATE \textbf{return} Estimated $\mathbf{p}$, $\boldsymbol{\theta}$, and $\mathbf{n}$
\end{algorithmic}
\end{algorithm}

\subsection{Key Challenges in IMU Sensors}
\label{subsec:challenges}
This study encountered several challenges when using a 9-axis IMU, which typically measures the 3-axis acceleration, angular velocity, and geomagnetic field. One of the key challenges is to obtain $\theta_i$. Traditionally, $\theta_i$ has been obtained from the 3-axis geomagnetic field measurements. However, when the IMU sensor is attached to a strut, magnetic interference becomes an issue, which leads to errors in $\theta_i$. This interference is primarily caused by the well-known hard and soft iron effects that are associated with ferromagnetic materials around IMU sensors \cite{ouyang2022analysis}, and these effects distort the magnetic field direction and cause errors in sensor measurements.

Additionally, using multiple IMU sensors in close proximity to one another further magnified the magnetic interference problem because the surrounding sensors influence each magnetometer sensor. To mitigate this issue, we opted to rely solely on the inclination angle $\phi_i$ from the 3-axis accelerometer (we also used the 3-axis angular velocity for sensor fusion), which effectively restricted the measurements to the 6-axis. This allowed us to obtain only stable inclination angles while using an optimization method (discussed in \hyperref[sec:ProposedAlgorithm]{Section II}) to estimate $\theta_i$. In this way, we can avoid complexities and errors from magnetic interference and provide a more stable and error-free estimated orientation.

Consequently, orientation  \( \mathbf {q}_i \) in (\ref{eq:q}) becomes the estimated orientation $\mathbf{\tilde{q}}_i\in \mathbb{R}^3$. This relationship is defined as follows.
\begin{equation} \tilde{\mathbf{q}}_i = f(\phi_i, \tilde{\theta}_i) \end{equation}
This redefined unit vector represents the estimated strut orientation, where $ \tilde{\theta}_i $ is obtained using the optimization method.

\subsection{Gradient Computation for Energy Minimization}
\subsubsection{Gradient for centre position of the struts}
To solve (\ref{eq:arg_min}), we used a numerical optimization method to efficiently solve the minimization problem of the total elastic energy $e$ with respect to changes in $\mathbf{p}$ using gradient information. The gradient is expressed as follows :
\begin{equation}
\frac{\partial e}{\partial \mathbf{p}} = \frac{\partial \mathbf{m_s}}{\partial \mathbf{p}}\left(\mathbf{K} \mathbf{m_s}\right)
\end{equation}
We use the iterative algorithm that updates $\mathbf{p}$ until the gradient satisfies  \( 
\left\|\frac{\partial e}{\partial \mathbf{p}}\right\| \approx 0,
\)
In other words, convergence occurs.
\subsubsection{Gradient for orientation of the struts}
We now compute the gradient of elastic energy, as defined in (\ref{eq:e_elongation}), with respect to \( \boldsymbol{\theta} \). This partial derivative, represented by \( \frac{\partial e}{\partial \boldsymbol{\theta}} \in \mathbb{R}^{1 \times m_b} \), can be written as a chain rule :
\begin{equation}
\begin{aligned}
& \frac{\partial e}{\partial \boldsymbol{\theta}} = \left(\frac{\partial e}{\partial \mathbf {m_s}} \right) ^T  \frac{\partial \mathbf {m_s}}{\partial \mathbf {q}} \cdot \frac{\partial \mathbf {q}}{\partial \boldsymbol{\theta}}
\end{aligned}
\label{eq:e/theta}
\end{equation}
The outer gradient of (\ref{eq:e/theta}) can be written as :
\begin{equation}
\begin{aligned}
& \frac{\partial e}{\partial \mathbf {m_s}} = \mathbf {K} \mathbf {m_s}
\end{aligned}
\label{eq:e/ms}
\end{equation}
The intermediate gradient of (\ref{eq:e/theta}) can be written by taking the partial derivative of $\mathbf {m_s}$ in (\ref{eq:ms}) with respect to $\mathbf {q}$ as :
\begin{equation}
\begin{aligned}
& \frac{\partial \mathbf {m_s}}{\partial \mathbf {q}} = \mathbf {C_s} \mathbf {B}^T 
\end{aligned}
\label{eq:8}
\end{equation}
The final gradient \( \frac{\partial e}{\partial \boldsymbol{\theta}} \in \mathbb{R}^{1 \times m_b} \) can be expressed as follows :
\begin{equation}
\frac{\partial e}{\partial \boldsymbol{\theta}} = \left(\mathbf {K} \mathbf {m_s} \right)^T  \mathbf {C_s} \mathbf {B}^T  \frac{\partial \mathbf {q}}{\partial \boldsymbol{\boldsymbol{\boldsymbol{\theta}}}}
\label{eq:de/dtheta}
\end{equation}

Likewise, we use an iterative algorithm that updates $\boldsymbol {\theta}$ until the gradient satisfies  \( 
\left\|\frac{\partial e}{\partial \boldsymbol {\theta}}\right\| \approx 0
\). 

 As discussed in \hyperref[subsec:challenges]{subsection H}, the complexity of obtaining $\theta_i$ by using a 3-axis magnetometer sensor. Therefore, we cannot compute the gradient 
 $\frac{\partial \mathbf {q}}{\partial \boldsymbol{\boldsymbol{\boldsymbol{\theta}}}}$, shown in (\ref{eq:de/dtheta}) directly from the magnetometer sensors. To mitigate this, we propose a method that estimates $\theta_i$ by minimizing the energy $e$. In the first step, we transformed the orientation from spherical to Cartesian coordinates shown in Fig.~\ref{fig:geometric}(b), the transformation can be written as :

\begin{equation}
\begin{aligned}
q_{ix} &= \sin \phi_i \cos \theta_i, & 
q_{iy} &= \sin \phi_i \sin \theta_i, & 
q_{iz} &= \cos \phi_i
\end{aligned}
\end{equation}

These equations describe how the components of \( \mathbf {q}_i \) in Cartesian coordinates (\( q_{ix}, q_{iy}, q_{iz} \)) are calculated based on \( \phi_i \) and \( \theta_i \) for strut \( i \). 
In (\ref{eq:de/dtheta}), each element of the gradient \( \frac{\partial \boldsymbol {q}}{\partial \boldsymbol {\theta}} \in \mathbb{R}^{3 m_b \times m_b} \) can be written as \( \frac{\partial \mathbf {q}_i}{\partial \theta_i} = \left[\frac{\partial q_{i x}}{\partial \theta_i}, \frac{\partial q_{i y}}{\partial \theta_i}, \frac{\partial q_{i z}}{\partial \theta_i}\right] \), and the partial derivatives of each component of \(\mathbf{q}_i\) can be written as follows :
\begin{equation}
\begin{aligned}
\frac{\partial q_{ix}}{\partial \theta_i} &= -\sin \phi_i \sin \theta_i, & 
\frac{\partial q_{iy}}{\partial \theta_i} &= \sin \phi_i \cos \theta_i, & 
\frac{\partial q_{iz}}{\partial \theta_i} &= 0
\end{aligned}
\end{equation}
These equations indicate that the \( x \) and \( y \)-components of \(\mathbf{q}\)  are functions of $\boldsymbol{\phi}$ and $\boldsymbol{\theta}$, whereas the \( z \)-component remains unaffected.

By computing the partial derivatives of the total energy \( e\) with respect to $\mathbf {p}$  and 
$\boldsymbol{\theta}$, we obtain the directions in which energy decreases most rapidly. We used a first-order optimizer, called Gradient Descent (GD) to iteratively adjust $\mathbf {p}$ and $\boldsymbol{\theta}$ in these directions, thereby converging towards the minimum energy shape of the structure.
\section{Experimental Results}
\label{sec:experimental}
This section provides a detailed description of the hardware setup used in the experiments and the implementation process. We then present the real-time shape estimation results under both static and dynamic conditions and discuss the findings.

 \subsection{Experiment Setup}
\begin{figure}[t]
    \centering  \includegraphics [width=86mm,height=36mm]
    {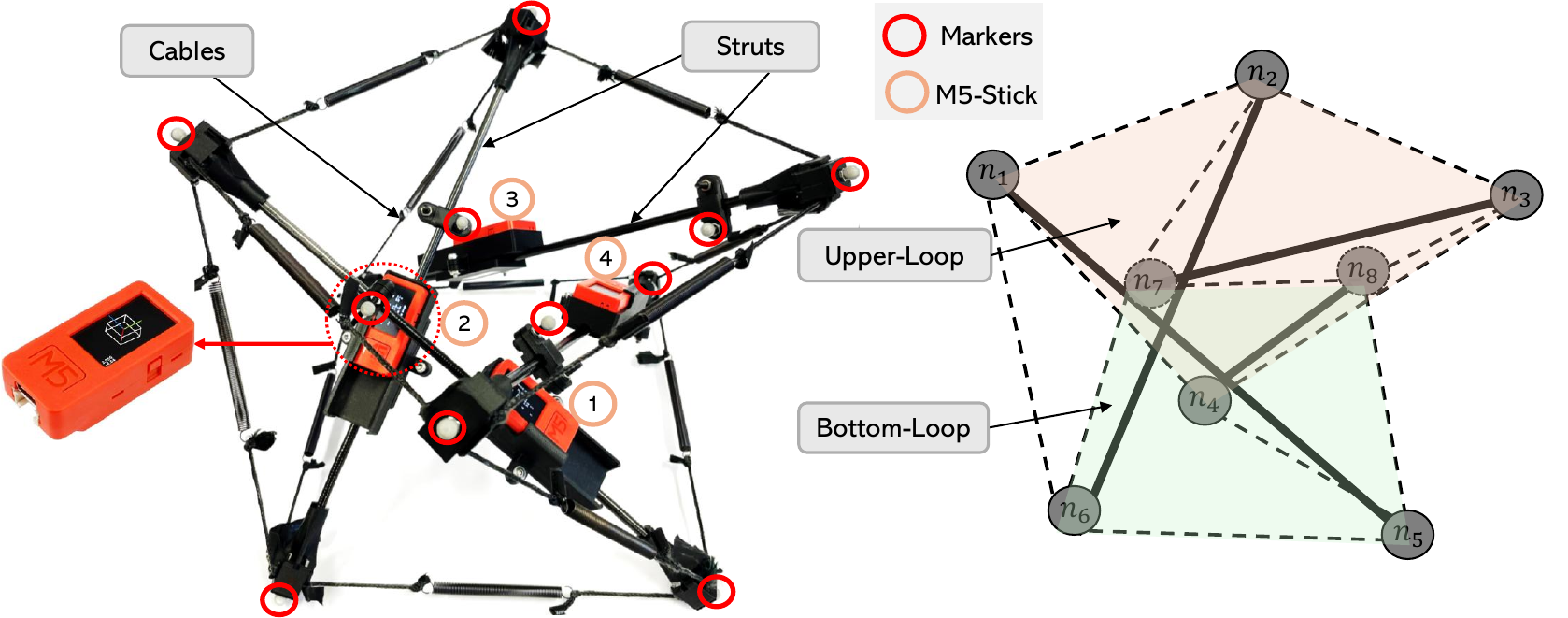}
    \caption{The experimental setup is based on a Class 1 tensegrity structure consisting of four struts and twelve cables. Each strut is equipped with an onboard M5stick device that includes an integrated IMU to measure the inclination angles $\boldsymbol{\phi}$. We also affixed three markers to each strut to obtain the ground truth.}
    \label{fig:class1}
\end{figure}

We conducted a series of experiments to evaluate the efficacy of our proposed algorithm. To do so, we implemented it on a Class 1 tensegrity structure that consists of four struts and twelve cables, as shown in Fig.~\ref{fig:class1}. Each cable measures 22~cm in length, and each strut has a length of 37~cm. We fabricated struts from carbon fiber-reinforced polymer (CFRP) because of its rigidity and lightweight properties. Each cable is created by connecting two strings at the end of a spring, with a spring constant of 0.064 N/mm. The tension in the cables is meticulously adjusted to maintain the structural shape and equilibrium. We place optical markers at the center of each strut and at the nodes, and we treat each strut as a rigid body that provides the ground-truth data. To determine $\boldsymbol{\phi}$, each strut is equipped with an M5StickC Plus 1.1 device, which consists of an ESP32-PICO-D4 dual-core microcontroller and a built-in IMU (MPU6886), which has a 3-axis gyroscope, a 3-axis accelerometer with an integrated Wi-Fi and Bluetooth module. The experiments are conducted in a controlled environment equipped with an optical motion capture set-up (OptiTrack PrimeX, utilizing eight infrared cameras) to directly measure the position and orientation of the struts, which are then used to obtain the ground truth of the actual system for the evaluation of the shape reconstruction. The Algorithm 1 uses the $\boldsymbol{\phi}$ obtained from IMUs and the \(\mathbf{L}\),  along with prior information of the \(\mathbf{C_s}\) and the \(\mathbf{K}\) of the cables, as inputs. The algorithm outputs (\ref{eq:nodalvector}), the spatial nodal positions, which have been used to reconstruct the shape of the structure.

\begin{figure*}[tb]
 \centering
     \includegraphics[width=\textwidth]{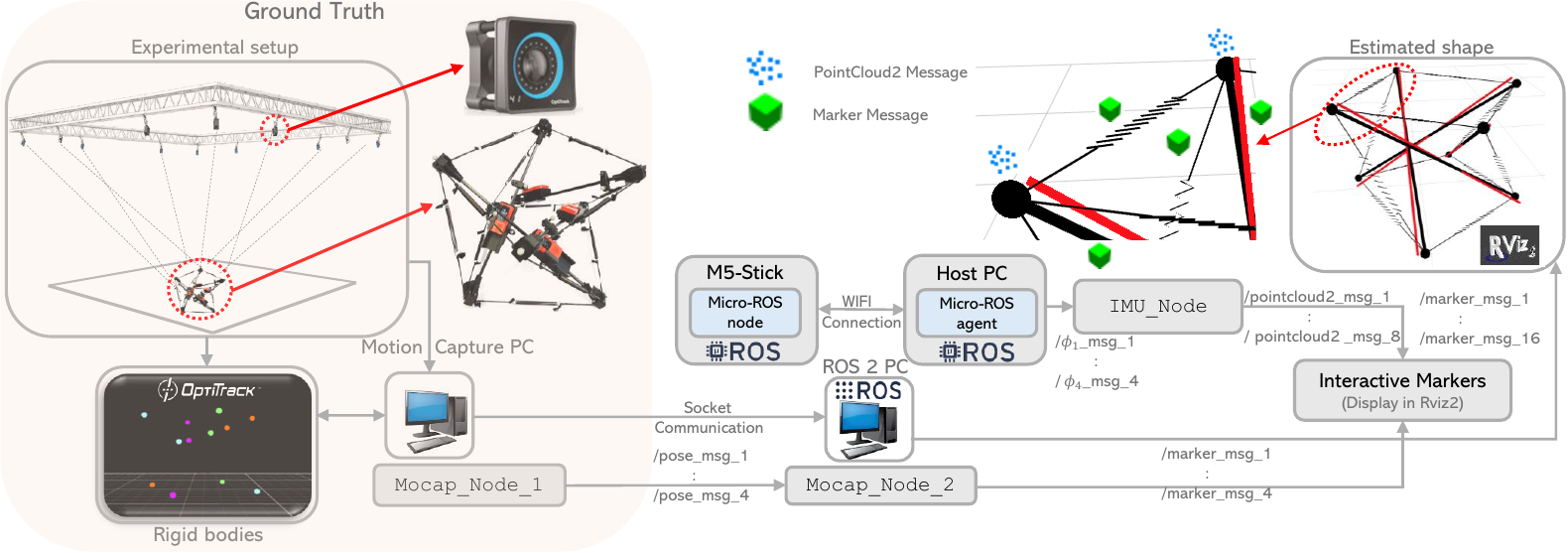}
     \caption{A high-level overview of the system architecture. The MoCap nodes are responsible for the ground truth (red) data published to RViz2, whereas the inclination angles $\boldsymbol{\phi}$ are used in the optimization process obtained from the M5-Stick. The optimized shape (black) has also been published to RViz2. For further details, refer to Section III.}
     \label{fig:whole_setup}
     \vspace{-1em}

\end{figure*}
\subsection{System Architecture}
The system architecture is shown in Fig.~\ref{fig:whole_setup}, we utilize a Motion Capture System (MoCap) with a Robot Operating System 2 (ROS 2) based control framework for real-time data acquisition and processing \cite{doi:10.1126/scirobotics.abm6074}. The MoCap system tracks the movement of each strut in a Class 1 tensegrity structure using multiple infrared cameras. It captures the position and orientation of the struts in real time and sends the data to the ROS 2 computer via a socket communication. A dedicated node, \texttt{Mocap\_Node\_1}, bridges the MoCap system and ROS 2 environment for efficient data transfer. The second node, \texttt{Mocap\_Node\_2}, subscribes to the incoming pose data and publishes them to the interactive marker module in RViz2. This module visualizes both shapes (estimated and actual) and allows real-time comparison between the estimated shape (black) and the MoCap-referenced ground truth (red). To visualize the estimated nodal positions, we utilized ROS 2 PointCloud2 messages.

In parallel, an \texttt{IMU\_Node} subscribes $\boldsymbol{\phi}$ from the M5-Stick, which functions as a Micro-ROS node \cite{belsare2023micro}. The M5-Stick runs the Micro-ROS node, and the host PC operates as the Micro-ROS agent. Data exchange between the node and the agent occurs over a Wi-Fi connection, which enables soft real-time communication with the ROS 2 network. The Micro-ROS node on the M5-Stick publishes $\boldsymbol{\phi}$ at 50 Hz from the IMUs, which are attached to each strut. The \texttt{IMU\_Node} subscribes to the $\boldsymbol{\phi}$ measurements for the optimization process. The optimization results are published in RViz2, visualized as part of the estimated shape, and overlaid with the MoCap ground truth for comparative analysis. The IMU data is continuously updated to maintain alignment with the ground truth. This parallel data processing ensures the synchronized handling of the MoCap and IMU streams, which ensures accurate real-time visualization. 
\subsection{Implementation}
\begin{figure}[H]
\centering
    \includegraphics[width=85mm,height=32mm]{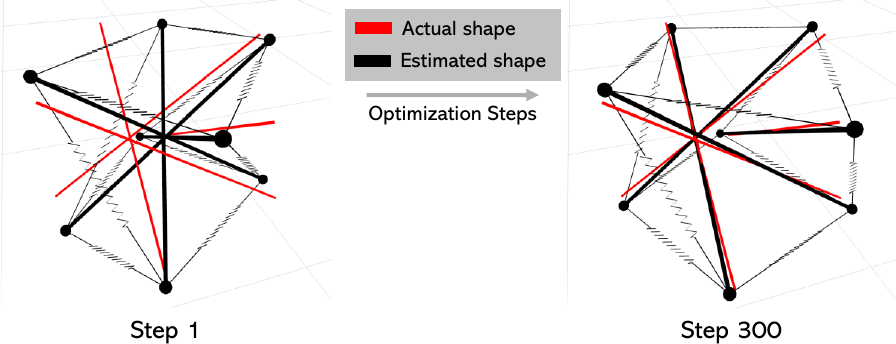}
    \caption{The figure shows the comparison of the actual shape (red) and the estimated shape (black) of the tensegrity structure at the initial step (Step 1) and after optimization (Step 300). The results show estimated shape improves and aligns more closely with the actual configuration as the optimization steps increase.}
    \label{fig:optimized}
    \end{figure}

The experiments are performed on a desktop computer with Ubuntu 22.04 LTS as the operating system and ROS 2 Humble as the software environment. We tested Algorithm 1 in Python, and on a system with an Intel Core i7-10870H processor. To evaluate the efficacy of our algorithm on a Class 1 tensegrity structure under static conditions, we used MoCap-referenced ground truth. As shown in Fig.~\ref{fig:optimized}, at the initial step (Step 1), the structure's estimated shape (black) is obtained using IMU sensors and the ground truth (red) from MoCap. Both shapes are visualized using ROS 2's RViz2 via marker messages. After 300 optimization steps, the estimated shape is approximately aligned with the ground truth and stabilizes in the static configuration, showing convergence.
\subsection{Dynamic Deformations}
\begin{figure}[tb]
 \centering
     \includegraphics[width=88mm,height=26mm]{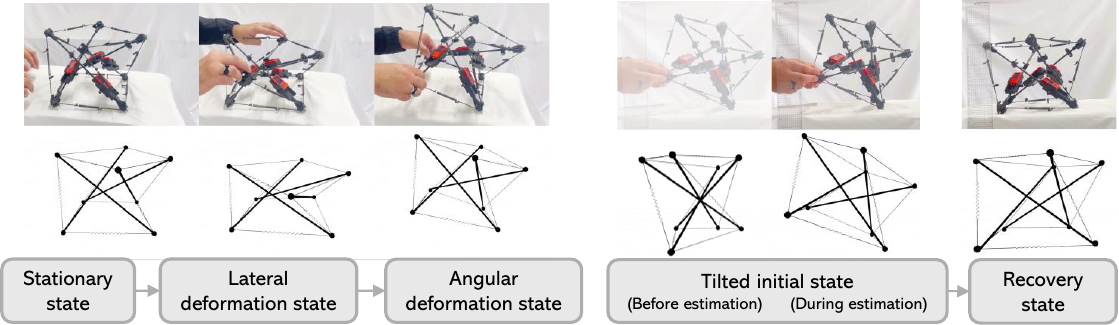}
     \caption{ The figure shows the different states of the structure: stationary, lateral deformation, angular deformation, tilted initial, and recovery states. The corresponding estimated shapes are shown below each state and visualized in RViz2. The deformations are manually induced to the structure.}
     \label{fig:states}
\end{figure}

\begin{figure}[tb]
 \centering
     \includegraphics[width=90mm,height=47mm]{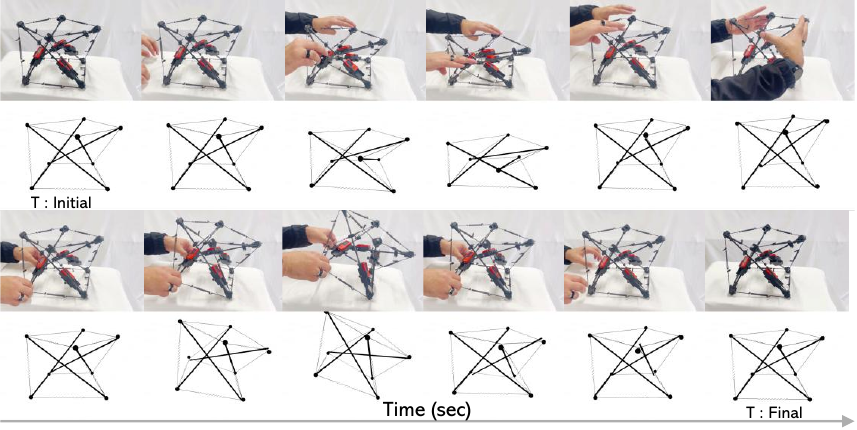}
     \caption{Real-time shape reconstruction of a Class 1 tensegrity structure during manual deformation using onboard IMU sensors. The figure shows a side-by-side comparison of the actual and the estimated shapes at different time instances.}
     \label{fig:frames}
\end{figure}

In the second experiment, the shape reconstruction is achieved through a sequence of states. Since our structure lacks actuators, we manually induced multiple deformations. The resulting deformations and their corresponding shapes are shown in Fig.~\ref{fig:states}.
\begin{itemize}
    \item {Stationary state:} In this state, the structure remains undeformed, and the estimated shape shows a static condition.
    \item {Lateral deformation state:} The structure is manually deformed perpendicularly, and the corresponding shape approximately reconstructs this deformed state.
    \item {Angular deformation state:} The structure is lifted from one side, similar to rolling locomotion \cite{kim2020rolling}, and it is also relevant to multiple tensegrity locomotion applications. The estimated shape also approximately matches the deformed state.
    \item {Tilted initial state:} We keep the structure initially on an inclined surface at an angle of $\sim$ 
 30 degrees from the horizontal plane. In the “During estimation” condition, the estimated shape shows the inclined deformed shape.
    \item {Recovery state:} After external deformation, the structure returns to its original state, and the estimated shape shows this recovery.
\end{itemize}
As shown in Fig.~\ref{fig:frames}, Algorithm 1 is tested on the tensegrity structure over a 30-second period. During this time, controlled external deformations are applied to assess the efficacy of the algorithm in real-time, and the results are visualized in RViz2. External deformations are induced at specific intervals, and the algorithm accurately estimates the resulting changes. 
\subsection{Results and Discussions}
\begin{figure}[tb]
\centering
\includegraphics[width=85mm,height=50mm]{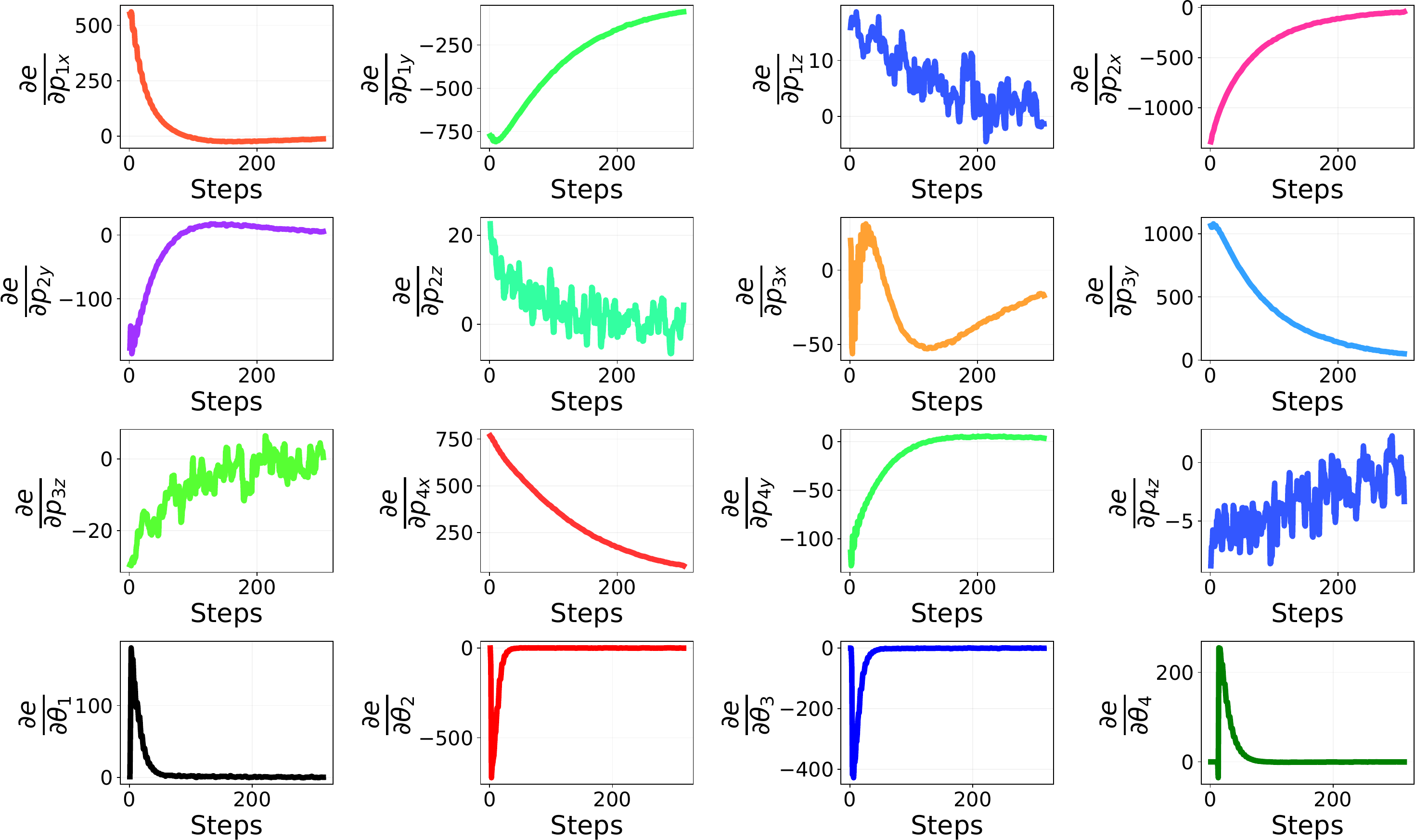}
\caption{The plots (top three rows) show the evolution of gradients of the energy function with respect to centre positions of the struts in three-dimensions, these include \( \frac{\partial e}{\partial p_{1x}}, \frac{\partial e}{\partial p_{1y}}, \frac{\partial e}{\partial p_{1z}} \)....... \( \frac{\partial e}{\partial p_{4z}} \). Similarly, the plots (fourth row) show the evolution of the energy gradients $\frac{\partial e}{\partial \theta_1}$, $\frac{\partial e}{\partial \theta_2}$, 
$\frac{\partial e}{\partial \theta_3}$, and $\frac{\partial e}{\partial \theta_4}$ 
over the optimization steps. Each plot shows the convergence of the optimization process. }
\label{fig:pp_gradient}  
\vspace{-1em}
\end{figure}

The experiments are conducted under both static and dynamic conditions. After each step, the centre positions and yaw angles of each strut are recalculated and updated using the GD method. This estimation process is repeated for a total of 300 steps.
\begin{figure}[t]
\centering
\includegraphics[width=78mm,height=33mm]{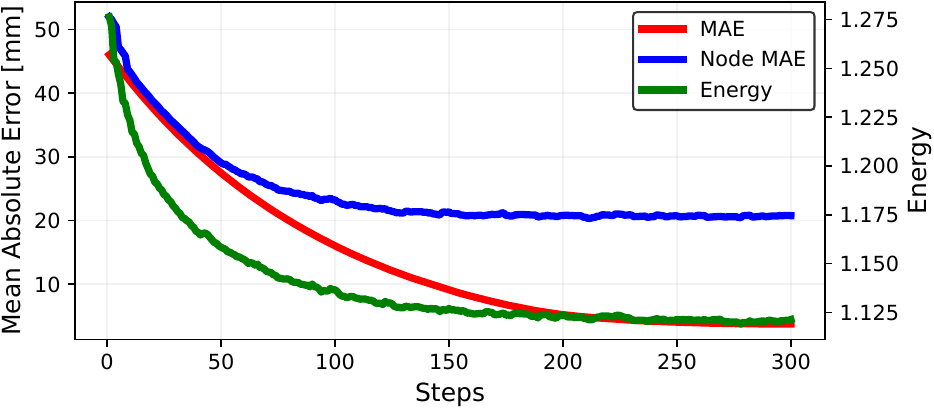}
\caption
  {The plot shows a monotonic decrease in the energy stored in the tensegrity structure over the optimization steps, showing convergence. Additionally, the plot shows the reduction of MAE for the strut center positions (red) and node positions (blue) during the optimization process.}
\label{fig:fig_combined}
\end{figure}Fig.~\ref{fig:fig_combined} shows a monotonic decrease in the total energy \( e\) of all cable elements, and the estimated shape converges to the true shape after approximately 300 steps. Additionally, Fig.~\ref{fig:fig_combined} shows the mean absolute error (MAE) between the actual and estimated shape for the center positions of the struts and nodal positions of the structure. Both errors decrease as the number of steps increases, eventually stabilizing at 3.76 mm and 20.78 mm,  even when we initialize the estimation values randomly. Fig.~\ref{fig:pp_gradient} shows the energy gradients for each strut with respect to their yaw angles and center positions in three-dimensional space. As the optimization approaches the optimal configuration, the gradients approach zero, satisfying the conditions \( 
\left\|\frac{\partial e}{\partial \mathbf{p}}\right\| \approx 0
\) and  \( 
\left\|\frac{\partial e}{\partial \boldsymbol {\theta}}\right\| \approx 0
\), and this shows we obtain an optimal configuration of the structure or a convergence. Table~\ref{tab:concatenated_angles_with_percentage_error} shows the estimated values of $\boldsymbol{\theta}$ after convergence is reached.
\begin{table}[h]
    \centering
    \caption{Percentage Errors Between Actual and Estimated Angles of Each Strut}
    \resizebox{\columnwidth}{!}{%
    \begin{tabular}{|c|c|c|c|}
        \hline
        Rigid Body & Actual Angle [rad] & Estimated Angle [rad] & Percentage Error (\%) \\
        \hline
        \multicolumn{4}{|c|}{\textbf{Yaw Angles (Angles Around Gravity Vector)}} \\
        \hline
        Strut 1 & $\theta_1 = 3.11$ & $\tilde{\theta}_1 = 2.94$ & 5.61\% \\
        \hline
        Strut 2 & $\theta_2 = 1.57$ & $\tilde{\theta}_2 = 1.52$ & 3.36\% \\
        \hline
        Strut 3 & $\theta_3 = 0.02$ & $\tilde{\theta}_3 = 0.16$ & 87.87\% \\
        \hline
        Strut 4 & $\theta_4 = -1.54$ & $\tilde{\theta}_4 = -1.49$ & 3.43\% \\
        \hline
        \hline
Rigid Body & Actual Angle [rad] & IMU Angle [rad] & Percentage Error (\%)\\  
\hline
        \multicolumn{4}{|c|}{\textbf{Inclinational Angles}} \\
        \hline
        Strut 1 & $\phi_1 = 0.95$ & $\tilde{\phi}_1 = 0.96$ & 1.47\% \\
        \hline
        Strut 2 & $\phi_2 = 0.96$ & $\tilde{\phi}_2 = 0.98$ & 1.87\% \\
        \hline
        Strut 3 & $\phi_3 = 0.95$ & $\tilde{\phi}_3 = 0.96$ & 0.52\% \\
        \hline
        Strut 4 & $\phi_4 = 0.96$ & $\tilde{\phi}_4 = 1.01$ & 5.78\% \\
        \hline
    \end{tabular}
    }
\label{tab:concatenated_angles_with_percentage_error}
\end{table}

To test the robustness, we performed 10 different trials using the GD method, and we determined that the algorithm converges to the same optimal solution with an average node MAE of 20.71 mm. We also conducted an ablation study in which we used the GD method as a baseline and compared its performance with that of the Stochastic Gradient Descent with Momentum (SGDM) and Adam optimizers. In total, we conducted 30 trials, with 10 trials for each optimizer.  The Table~\ref{tab:diffopt} shows that all three methods converge to approximately the same optimal solution, with Adam performing better in terms of convergence time. We also computed the convergence time, which is defined as the average runtime from the random initialization of both $\mathbf{p}$ and $\boldsymbol{\theta}$ to reach the thresholds $\left\|\frac{\partial e}{\partial \mathbf{p}}\right\| \approx 0$ {and} $\left\|\frac{\partial e}{\partial \boldsymbol{\theta}}\right\| \approx 0$, respectively. And for each update step, the algorithm takes $\sim$  0.52 \textit{ms}. The computed time in Table~\ref{tab:diffopt} shows that the proposed method operates at high frequencies and can be employed for real-time dynamic shape estimation.
\begin{table}[H]
\caption{Performance Comparison of Different Optimizers\label{tab:optimizer_comparison}}
\centering
\setlength{\tabcolsep}{1.5pt} 
\renewcommand{\arraystretch}{1.05} 
    \resizebox{1\columnwidth}{0.07\columnwidth}{

\begin{tabular}{|c|c|c|c|c|c|}
\hline
Optimizer & Nodal MAE (Mean \(\pm\) Std)
 & Energy (Mean \(\pm\) Std) & Step Time (\textit{ms})   & Avg. Time (\textit{ms})  \\
\hline

GD & 20.71 \(\pm\) 0.15 & 1.12 \(\pm\) 0.00 & 0.52  & 97.16  \\
\hline
SGDM & 21.10 \(\pm\) 0.45 & 1.12 \(\pm\) 0.00  & 0.52  & 80.76 \\
\hline
Adam & 20.99 \(\pm\) 0.26 & 1.12 \(\pm\) 0.00 & 0.60 & 67.24  \\
\hline
\end{tabular}}
\label{tab:diffopt}
\vspace{-1em}
\end{table}
At the 300\textsuperscript{th}
 steps of the optimization, as shown in Fig. ~\ref{fig:optimized} the estimated shape does not completely overlap the actual shape. Several factors are likely to contribute to this discrepancy, including the constant error between the inclination angles of the struts obtained from the IMU sensors and the actual inclination angles listed in Table~\ref{tab:concatenated_angles_with_percentage_error}. 
 
 In addition, we used the optimized stiffness matrix $\mathbf{K}$, obtained using the Optuna framework \cite{akiba2019optuna}, which is an automated tool for hyperparameter optimization. These values are used instead of the actual spring constants, which may require further tuning to improve accuracy. Finally, we created a comparison Table~\ref{tab:sensor_comparison} that shows our approach is based on onboard sensors and requires fewer types of sensors (one) compared to existing methods for real-time shape reconstruction.

\renewcommand{\arraystretch}{0.8}

\section{Conclusion and future work}
\label{sec:conclusion}
In this study, we present a novel approach to estimate the shape of a Class 1 tensegrity structure in real time using only the inclination angle information of the struts. To validate its efficacy, we implemented the algorithm on a Class 1 tensegrity structure with onboard IMU sensors. The algorithm successfully estimated the shape of the structure under static conditions and when subjected to external deformation. Our method achieved an MAE of 3.76 mm and 20.78 mm for the strut centers and node positions, respectively. This shows that our algorithm provides reliable real-time shape estimation and could be suitable for many real-world tensegrity applications. We believe that this algorithm is applicable to tensegrity structures with minimal hardware modifications.
\begin{table}[H]
\centering
\renewcommand{\arraystretch}{1.1}
\caption{ Comparison of Shape Reconstruction Approaches}
\label{tab:sensor_comparison}\

\begin{tabular}{|p{0.6cm}|p{1.3cm}|p{0.8cm}|p{1.6cm}|p{2.3cm}|}
\hline
Ref. & Sensor  \newline Dependency & Shape \newline Reconst. & Approach & Sensor Types\\ 
\hline
 \cite{tong2025tensegrity} & Onboard  & 3D & Constrained \newline Optimization & IMUs, Encoders \\
\hline
\cite{caluwaerts2016state} & Onb/Ex & 3D & Unscented Kalman Filter & IMUs, Encoders,\newline Ranging Sensors\\
\hline
\cite{bezawada2022shape}  & Onb/Ex & 2D & Energy Minimization & IMUs, Vision-based \newline Markers \\
\hline
\cite{lu20226n} & Onb/Ex & 3D & Iterative Optimization & Elastic Stretch, \newline RGB-D camera   \\  
\hline
Our & Onboard & 3D & Energy Minimization & IMUs \\ 
\hline
\end{tabular}
\end{table}

In future work, we plan to extend this approach by implementing the algorithm on our existing tensegrity robot TM40 \cite{yoshimitsu2022development}, which consists of five Class 1 tensegrity structures stacked vertically, as shown in Fig.~\ref{fig:TM40 tensegrity manipulator}. Unlike single-layer Class 1 tensegrity structures, this system does not rely on mechanical springs. Instead, all deformations are induced by active and passive displacements of the pneumatic cylinders. These pneumatic cylinders behave like mechanical springs, but their stiffness varies dynamically in response to changes in internal pressure.

\section*{Acknowledgments}
We sincerely thank Mr. Murai, a former member of our lab, during his master’s program for his contribution in developing the foundational concepts of this algorithm.

\bibliographystyle{IEEEtran}
\bibliography{references} 

\end{document}